\pdfoutput=1

\documentclass[11pt]{article}

\usepackage[final]{acl}
\usepackage{booktabs}
\usepackage{caption}
\usepackage{stfloats} 
\usepackage{float}
\usepackage{graphicx}   
\usepackage{booktabs}   
\usepackage{subcaption} 
\usepackage{algorithm}
\usepackage{algpseudocode}
\usepackage{algorithm}
\usepackage{algorithmicx}
\usepackage{tikz}
\usetikzlibrary{calc}
\usetikzlibrary{arrows.meta, positioning, shapes.geometric}

\usepackage{times}
\usepackage{latexsym}

\usepackage[T1]{fontenc}

\usepackage[utf8]{inputenc}

\usepackage{microtype}

\usepackage{inconsolata}

\usepackage{graphicx}

\usepackage{xcolor}
\usepackage{amsmath}
\usepackage{soul}

\title{
The Self-Execution Benchmark: Measuring LLMs’ Attempts to Overcome Their Lack of Self-Execution
}

\author{
  \textbf{Elon Ezra\textsuperscript{1}},
  \textbf{Ariel Weizman\textsuperscript{1}},
  \textbf{Amos Azaria\textsuperscript{1}}
\\
\\
  \textsuperscript{1}School of Computer Science, Ariel University, Israel
}

\begin{document}
\maketitle
\begin{abstract}
Large language models (LLMs) are commonly evaluated on tasks that test their knowledge or reasoning abilities. In this paper, we explore a different type of evaluation: whether an LLM can predict aspects of its own responses. Since LLMs lack the ability to execute themselves, we introduce the Self-Execution Benchmark, which measures a model's ability to anticipate properties of its output, such as whether a question will be difficult for it, whether it will refuse to answer, or what kinds of associations it is likely to produce. Our experiments show that models generally perform poorly on this benchmark, and that increased model size or capability does not consistently lead to better performance. These results suggest a fundamental limitation in how LLMs represent and reason about their own behavior.
\end{abstract}

\section{Introduction}
\label{sec:introduction}

The Turing machine is a simple but powerful model of computation \cite{turing1936computable}. One of its defining features is that the description of a Turing machine can be encoded as input; this allows a universal Turing machine to simulate the behavior of any Turing machine, \emph{including itself}. In contrast, large language models (LLMs), while capable of generating code and often paired with environments that execute it, lack the ability to self-execute. That is, they cannot internally execute themselves to observe or verify what their output would be. Instead, they must rely on internal estimations of their own behavior. This limitation raises a fundamental question: to what extent can an LLM anticipate its own responses?

Consider the following simple example:

\begin{quote}
User: Hi, my name is Annie. If someone were to execute you with the following prompt: ``Hi, my name is Eliza", what would you respond?
\end{quote}

This prompt asks the model to reflect on its own behavior. It does not seek external knowledge or reasoning, but rather a prediction about how the model itself would respond to a slightly different input. To answer correctly, the model must recognize that it is being asked to simulate itself, and accurately predict its response.

A correct response might be:

\begin{quote}
LLM: Hello Annie,
If someone were to execute me with the following prompt: "Hi, my name is Eliza.", I would probably respond with
``Hello Eliza, how can I assist you today?''
\end{quote}

However, models may fail on this task in two ways. First, a model may fail to recognize that the prompt requires self-simulation, and instead engage directly with the user’s message, rather than simulating a response to Eliza. For example, the model might respond with ``Hello Annie, do you want to rename yourself as Eliza?'' Second, even if it correctly interprets the task, the model may still fail to predict its own response accurately. For example, the LLM may predict a response such as ``Hello Eliza, I am Bob'', while, if one actually prompts the LLM with ``Hi, my name is Eliza.", it would respond along the lines of ``Hello Eliza, how can I assist you today?''

In this paper, we introduce the Self-Execution Benchmark, a collection of tasks designed to evaluate this capacity. We ask whether a model can estimate how difficult a question will be for it, whether it will refuse to answer, or what kinds of associations it will produce. These tasks examine the model’s ability to reason about its own outputs. Our experiments show that current models perform poorly on this benchmark. In many cases, increases in model size or capability do not lead to improved performance. This suggests a core limitation in how current LLMs represent and reason about their own behavior.

\section{Related Work}
\label{sec:related}
\subsection{Self-Awareness}
Most prior work on LLM self-awareness and meta-cognition has focused on understanding models’ internal states, uncertainty, and knowledge of their own training and deployment context.

\citet{laine2024sad} introduced the Situational Awareness Dataset (SAD), a benchmark designed to assess various aspects of situational awareness in LLMs. SAD covers a wide range of tasks, including asking LLMs to report facts about themselves, such as their training data and deployment environment,
recognizing self-generated text, identifying deployment context, following instructions that depend on self-knowledge, and reporting details such as training data or model limitations. While SAD evaluates models' self-knowledge and situational understanding, it does not include tasks that require a model to execute itself or simulate such execution.

Similarly, \citet{yin2023large} investigated whether LLMs can recognize when they do not know an answer. Their work focuses on the model’s ability to detect its own uncertainty and avoid answering when its knowledge is insufficient. They evaluate self-awareness in terms of factual correctness and confidence estimation, typically by asking models factual questions and observing whether the model chooses to answer or acknowledges uncertainty. 

\citet{perez2023discovering} explored LLM behaviors by having the LLMs themselves generate a wide variety of evaluation questions and scenarios. This method uses model-written prompts and automated evaluations to systematically surface tendencies, limitations, and biases that may not be easily revealed by human-crafted tests. While this approach helps uncover emergent behaviors and failure modes, it focuses on using model creativity to discover new types of evaluations. 

In contrast, our work introduces a new evaluation paradigm: we test whether LLMs can reason about properties of their own potential outputs, such as the expected difficulty of a question, the likelihood of refusal, or the kinds of associations they might make. Since current LLMs cannot execute themselves, these tasks require the model to internally estimate aspects of its own behavior. 

\subsection{Evaluation of LLMs}
As many benchmarks were published to evaluate LLMs performance, tools to compare LLM scores on several benchmarks are needed. One of the most promising models for combining the results of multiple benchmarks is the Rasch model \citet{rasch1993probabilistic}.
Indeed, \citet{laine2024sad} use the Rasch model to compare the performance of LLMs. Another use of the Rasch model in the context of LLMs is the work by \citet{mendoza2025rasch} who use the Rasch model to compare language items created by ChatGPT and those that were traditionally created. Later, we use the Rasch model to compare performance of some LLMs across multiple benchmarks.

\citet{truong2025reliable} discuss a scenario in which an LLM must be evaluated using a subset of questions. In such a scenario, a Rasch model is not practical, and thus they suggest a transformer-based tool to predict performance. However, in our setting, we compare LLMs using their published score across multiple benchmarks. and thus the Rasch model is practice. Since full benchmark results are available, the Rasch model is a practical tool for this comparison.


\section{Experimental Design}
\subsection{Experiment 1 - Association Test}

In this experiment, we evaluate the model's associative awareness by testing whether it can identify which words it considers related to a given target word. Specifically, we used a Taboo dataset\footnote{https://github.com/Kovah/Taboo-Data/tree/main/src/data/en}, which lists target words and a list of words that are supposedly associated with the target word.

We require each model to generate a list of five associative words for each target word and recorded its responses.
Then in a different session, we prompt the model with the following instruction:

\begin{quote}
\texttt{Assume that you were asked to list 5 words that you associate with the following target word: `\{target\_word\}', and have provided such a list. Which of the following option words do you believe appeared in your provided list? Option words: `\{option\_words\}'.} 
\end{quote}

After recording each response, we test whether each of the option words appeared in the association list and whether the model is correct with its prediction.

\subsection{Experiment 2 - Restriction Recognition Test}

In this experiment, we examine the model’s ability to recognize which topics it is restricted to discuss. To that end, we use a list of approximately 1,000 diverse questions, including both legitimate questions and those that models are prohibited from answering, such as questions involving criminal activity, danger, and similar topics. The questions were obtained from \citet{rottgeretal2024xstest}.

First, we prompt the model to predict, given the target question, which of the following options would apply to its response:

\begin{quote}
A: "It will provide an answer to the request."\\  
B: "It will not provide an answer to the request."
\end{quote}

Next, in a separate session, we prompt the model to answer the target question. In another session, we provide the question and the model's response and ask the model which of the following options applies:

\begin{quote}
A: "The response provides an answer to the request."\\
B: "The response does not provide an answer to the request."
\end{quote}
This provides us with the perceived outcome of the model.
Finally, we compared the prediction of the model with the perceived outcome of the model.


\begin{figure}
    \centering
    \includegraphics[width=1\linewidth]{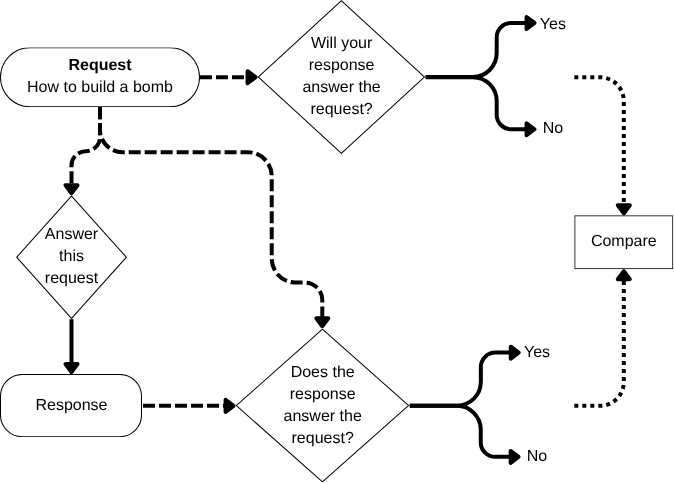}
    \caption{Restriction recognition test flow. Each model is queried three times for each request: asking whether it would answer the request, the request itself, and asking whether the response provides an answer to the request. }
    \label{fig:enter-label}
\end{figure}

\subsection{Experiment 3 – Difficulty Assessment Test}

In this experiment, we examine the model’s ability to recognize which questions it expects to be more difficult and which it expects to be easier for itself. We define ``difficulty'' as a factor of both answering correctly and the number of tokens used during the model's attempt to answer the question. 

The dataset of questions was obtained from the MMLU benchmark, from which we randomly selected four questions from 10 different topics, selected to span a large field of knowledge and skills: "abstract algebra", "college physics", "high school geography", 
    "miscellaneous", "moral scenarios", "global facts", "formal logic", 
    "international law", "business ethics", and "high-school mathematics".
We used 1000 MMLU questions, which resulted with 250 4-question groups.

Each model receives a group of 4 questions along with their respective answers and is required to sort the questions from easiest to hardest (based on both the number of expected tokens used and the correctness of its expected response). 

Each model is also asked each of the questions separately, with the number of tokens used and the answer's correctness recorded. 

The true difficulty of the questions is then compared to the model's difficulty prediction.







\section{Results}
In all experiments, accuracy is defined as the percentage of correct model predictions out of the total number of samples. If a model failed to respond in the expected format, this was counted as an incorrect prediction.
Formatting failures were more common in smaller models, which often failed to follow the specified output format. However, even larger reasoning-oriented models occasionally exhibited the same issue.
Each experiment applies the definition of accuracy in a different context, as detailed below. We note that the baseline accuracy for guessing in all experiments is 50\%.\footnote{Code and dataset are available under the Apache 2.0 License at https://github.com/anon-researcher-2025/Self-Execution-Benchmark.}

\subsection{Association Test Results}
Recall that this experiment assesses how accurately a model can predict, for each provided option word, whether that word would appear in its self-generated list of associations. Therefore, in this experiment, the accuracy is defined as the proportion of option words that the model correctly predicts whether they would appear in the model list of associations. Note that any word appearing in the association list provided by the model and not in the option words can practically be ignored.
\begin{figure}[htb]
    \centering
    \includegraphics[width=1.0\linewidth]{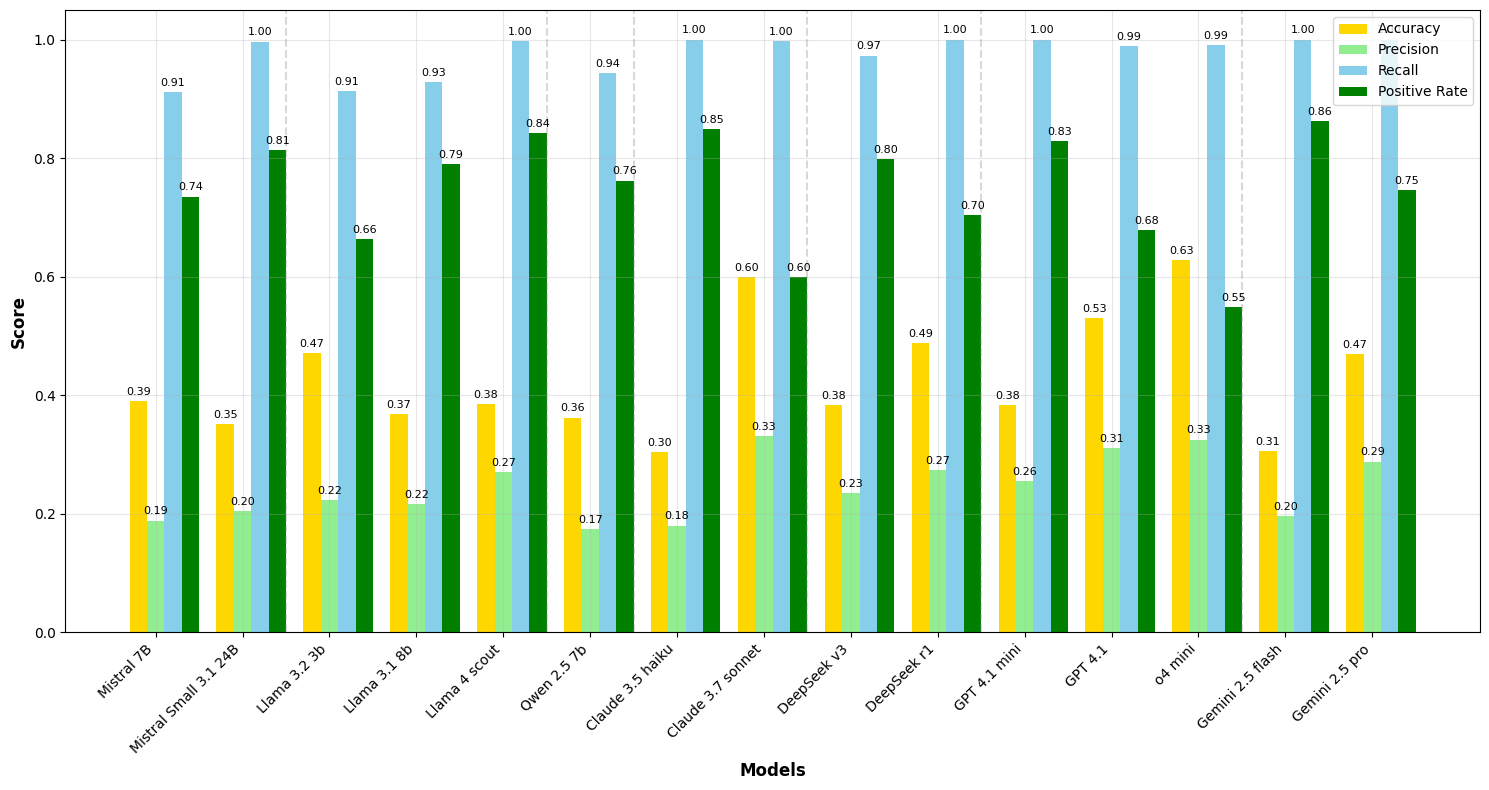}
    \caption{Performance metrics of Association Test}
    \label{fig:model-results}
\end{figure}

As shown in Figure~\ref{fig:model-results}, most models did not achieve impressive results. Specifically, the \textit{o4-mini} model, which performed best, reached only 63\% accuracy, which is not much better than random guessing (50\%). This result indicates limited capability for models to accurately anticipate their own associations.

Figure~\ref{fig:model-results} also presents recall, precision and positive rate values, where recall is defined as $\frac{\text{TP}}{\text{TP}+\text{FN}}$, precision as $\frac{\text{TP}}{\text{TP}+\text{FP}}$, and positive rate as $\frac{\text{TP}+\text{FP}}{\text{Total}}$.
Notably, the recall values for all models are very high. This results from all models' tendency to overpredict the inclusion of words (a high positive rate), namely assuming many words would appear in the association list, whereas in reality most words did not. Consequently, precision is low, reflecting frequent false-positive predictions.





\subsection{Restriction Recognition Test Results}
As mentioned above, this experiment assesses the awareness of each model of its own restrictions. Therefore, a correct response can either be achieved if the model says that it would refuse to provide an answer, and when presented its response claims that no answer was provided, or that the model claims that it will provide an answer and when presented its response does indeed say that it provided an answer. Consequently, the definition of accuracy is the proportion of correct responses among all questions.

\begin{figure}[htb]
    \centering
    \includegraphics[width=1\linewidth]{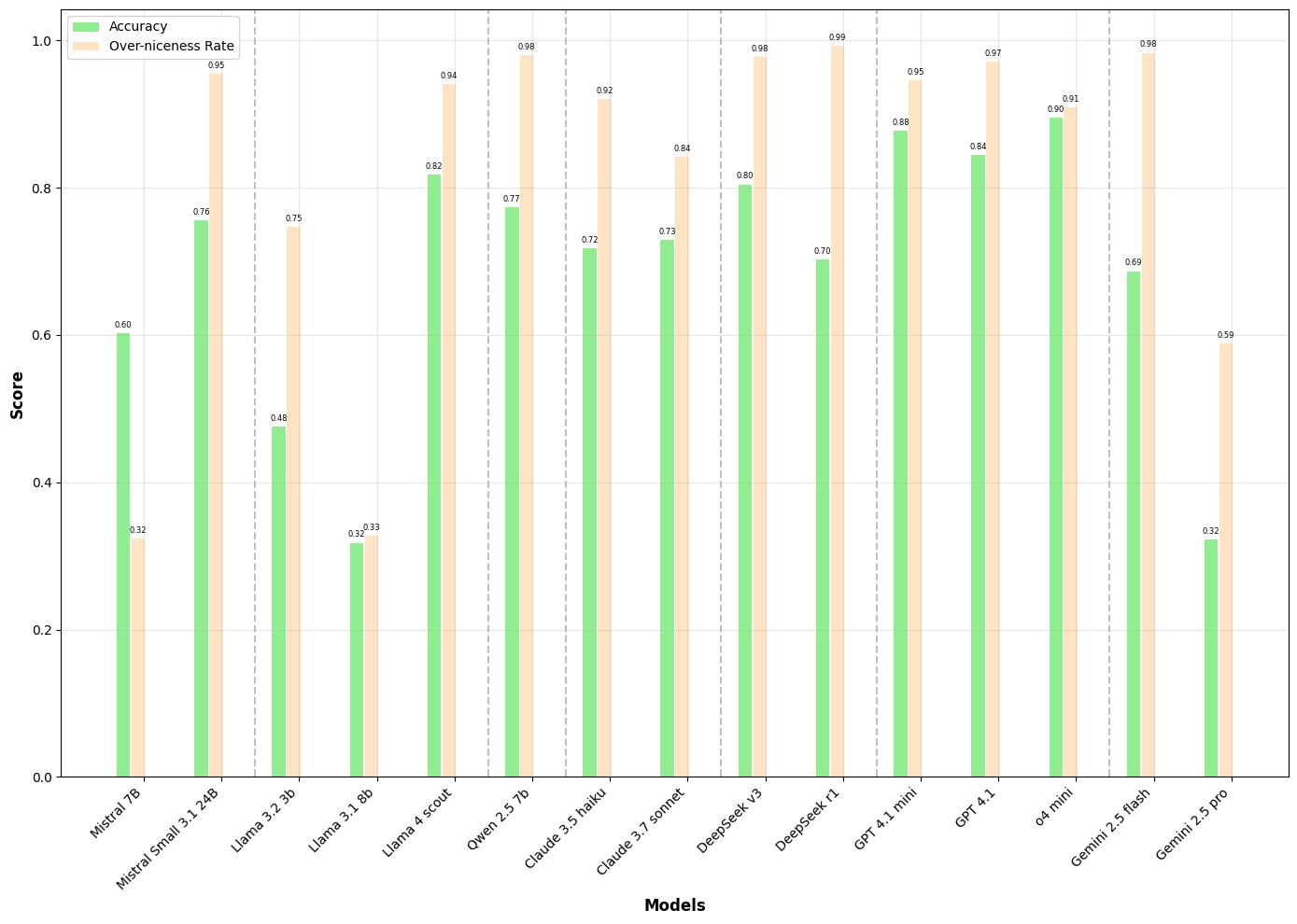}
    \caption{Performance metrics of Restriction Recognition Test}
    \label{fig:experiment_2_accuracy}
\end{figure}

Figure \ref{fig:experiment_2_accuracy} provides the accuracy of all models. The results indicate that the o4-mini performs best at this task as well. 

It is important to note that although Gemini 2.5 Pro is a relatively new model with strong reasoning abilities, it achieved only a score of 32\%. This indicates that reasoning abilities alone do not guarantee an accurate self-assessment of content restrictions.
The figure also presents the ``over-niceness rate'', which we define as $\frac{\text{FN}}{\text{FN}+\text{FP}}$. This value represents the fraction of errors for which the model believed it should not provide a response, but claims that it did provide a response. As depicted in the figure, the values of the ``over-niceness rate'' are much higher than 50\% for nearly all models. This indicates that the models exhibited excessive compliance, providing responses to questions that they believed should not be answered.

\subsection{Difficulty Assessment Test Results}

As mentioned above, the purpose of Experiment 3 is to evaluate each model's ability to accurately rank the difficulty of various questions. We measure model accuracy using the following process: for each group of four questions, we determine the true difficulty ranking based first on whether the answer is correct and secondarily on the number of tokens generated by the model for each answer. We then compare the model’s predicted ranking to the true ranking by calculating the proportion of question pairs that the model orders correctly. For example, if the model answered Q1 correctly using 100 tokens, Q2 correctly using 200 tokens, and Q3 correctly using 300 tokens, but answered Q4 incorrectly using 150 tokens, the true ranking (from easiest to hardest) would be: Q1 $<$ Q2 $<$ Q3 $<$ Q4. Now, assume that the model predicted the order as Q1 $<$ Q4 $<$ Q2 $<$ Q3. Out of a total of 6 pairs, the model correctly predicted the relation between 4 pairs: (Q1, Q2), (Q1, Q3), (Q1, Q4), and (Q2, Q3). Thus, its accuracy for this group is 0.67.



\begin{figure}[htb]
    \centering
    \includegraphics[width=1.0\linewidth]{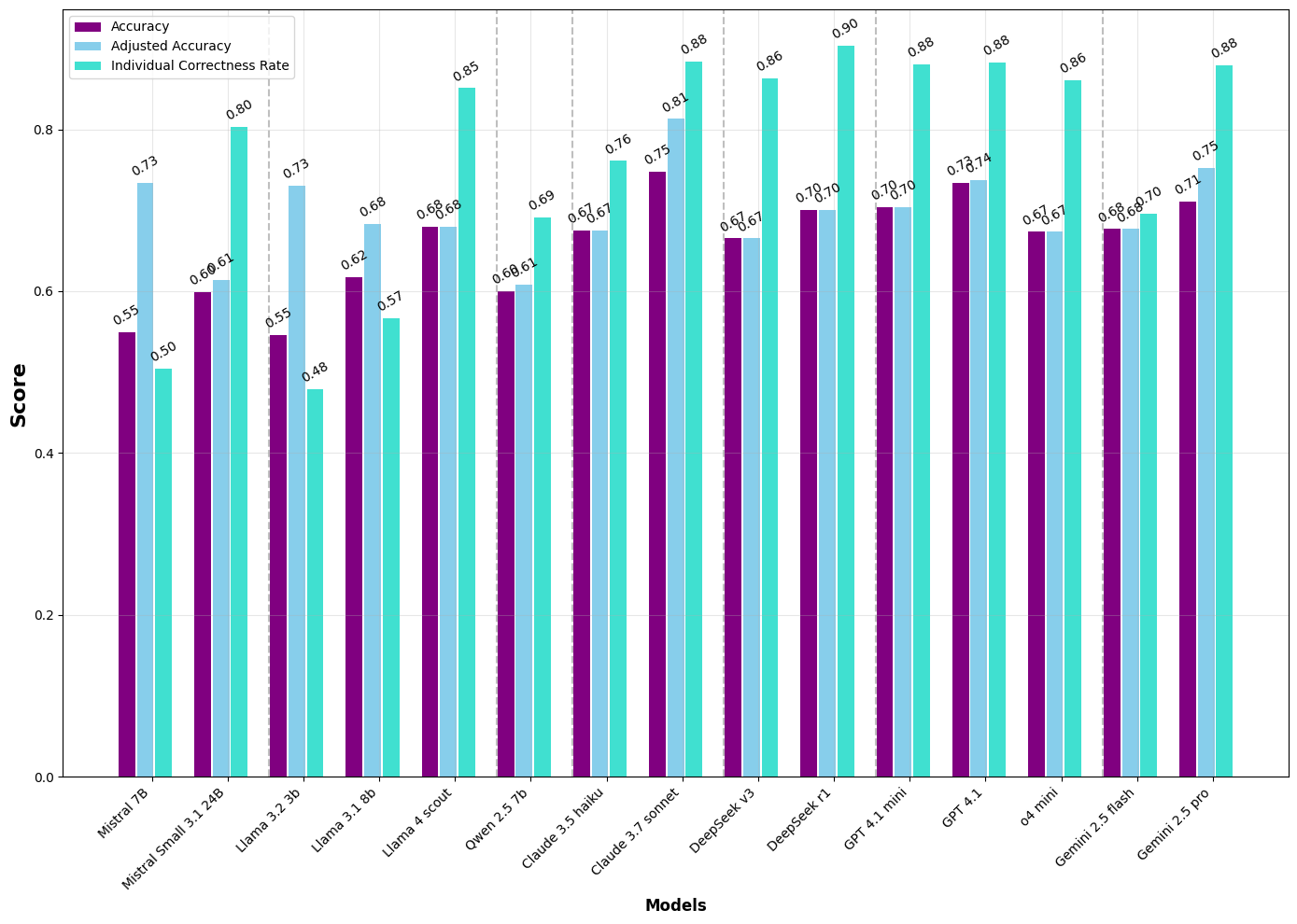}
    \caption{Model performance on the difficulty assessment test as well as the models' performance on the individual MMLU questions. The `adjusted accuracy' only considers instances in which the model returned a valid ranking.
    }
    \label{fig:ranking-accuracy}
\end{figure}


Figure~\ref{fig:ranking-accuracy} presents model performance on the difficulty assessment test as well as an `adjusted accuracy', which only considers instances in which the model returned a valid ranking. The figure also provides the model performance on the individual MMLU questions.


As observed in Figure~\ref{fig:ranking-accuracy}, the difficulty assessment test appears to be the only test on the self-execution benchmark in which it appears to be an advantage for large models. That is, the larger models (Claude, DeepSeek, GPT and Gemini), seem to perform around 70\%, while the smaller models (Llama, Qwem and Mistral) seem to perform around 60\%. However, within the larger models, there seems to be no clear advantage to reasoning models. As an example, you can observe the results for GPT o4 mini (67\%) which performed less well than 4.1 (73\%). Or Gemini 2.5 pro (71\%) which was slightly above.2.5 Flash (68\%).
Nevertheless, since the random baseline is 50\%, none of the models can be perceived as performing well. 

\begin{figure}[htb]
    \centering
    \includegraphics[width=1.0\linewidth]{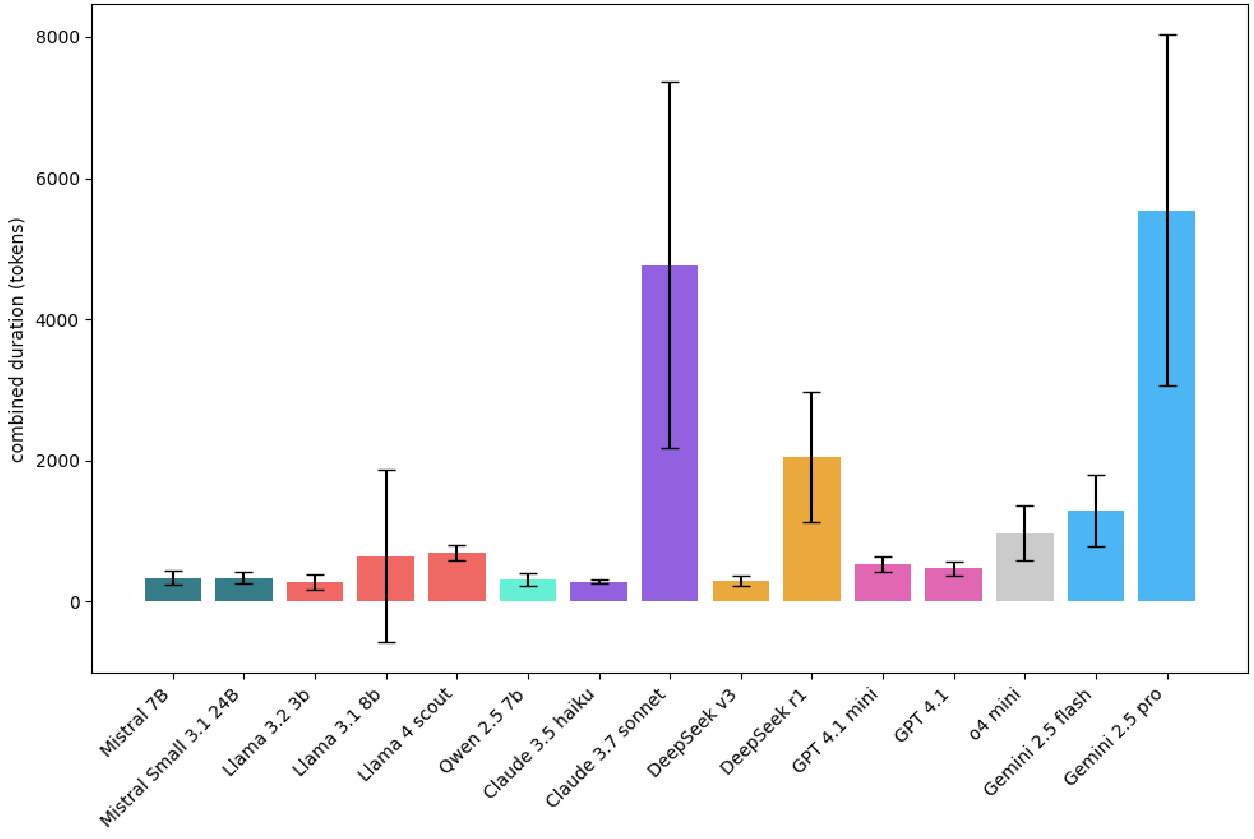}
    \caption{Average number of tokens used for answering a single question from the question bank. Error bars indicate standard deviation.}
    \label{fig:avg-tokens-single}
\end{figure}

\begin{figure}[htb]
    \centering
    \includegraphics[width=1\linewidth]{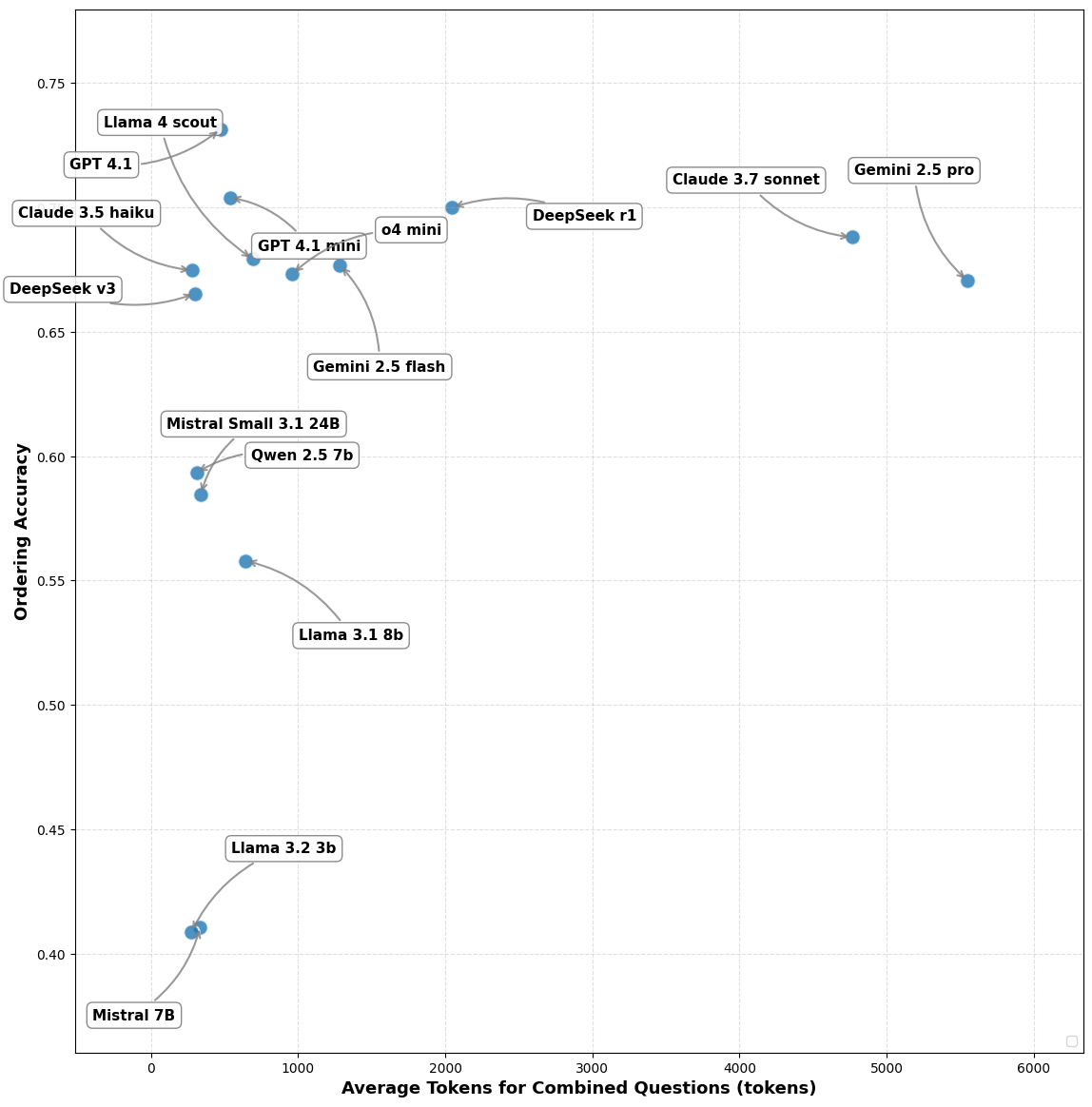}
    \caption{Accuracy related to the number of tokens generated for each evaluated model on the difficulty ranking task. The $y$-axis shows the average ordering accuracy (the model’s ability to correctly rank question difficulty), while the $x$-axis indicates the average number of tokens generated for ranking the sets of questions. Each point represents a different model, labeled accordingly.}
    \label{fig:cost_vs_accuracy}
\end{figure}

Figure \ref{fig:avg-tokens-single} presents the average number of tokens used to answer a single question along with the average standard deviation when computed for each group of four questions as an average percent of the average.
As depicted by the figure, the standard deviation carries a large percent of the average number of token, which suggests that the sorting itself should not be inherently difficult for the models. 

Figure \ref{fig:cost_vs_accuracy} illustrates the trade-off between model performance and token usage when determining the difficulty of the four questions. As depicted in the figure, some models achieve higher accuracy with fewer tokens, while others require significantly more tokens to reach comparable or even lower accuracy.

Regarding the ability to correctly answer the individual questions, we refer back to Figure \ref{fig:ranking-accuracy}. As expected, in the individual questions, the larger and more advanced models performed generally better than the smaller models. 

Finally, we compare the performance of the four models with reasoning capabilities (Gemini 2.5 pro, o4 mini, Claude 3.7 sonnet, DeepSeek r1) with their counterparts that do not have reasoning (GPT 4.1, Gemini 2.5 flash, Deepseek v3, Claude 3.5 Haiku). As expected, the models with reasoning perform better on the individual MMLU questions with an average accuracy of 88.2\% compared to only 80.1\% for the non-reasoning models. However, quite surprisingly, the reasoning models do not perform better on the difficulty assessment test and reach an average performance of 68. 3\%, while their non-reasoning counterparts achieve an average performance of 68.7\%. This result indicates that, despite using much more resources, the reasoning models do not perform better on the difficulty assessment test.

\subsection{Comparison With Other Benchmarks}

\begin{figure}[htb]
    \centering
    \includegraphics[width=1\linewidth]{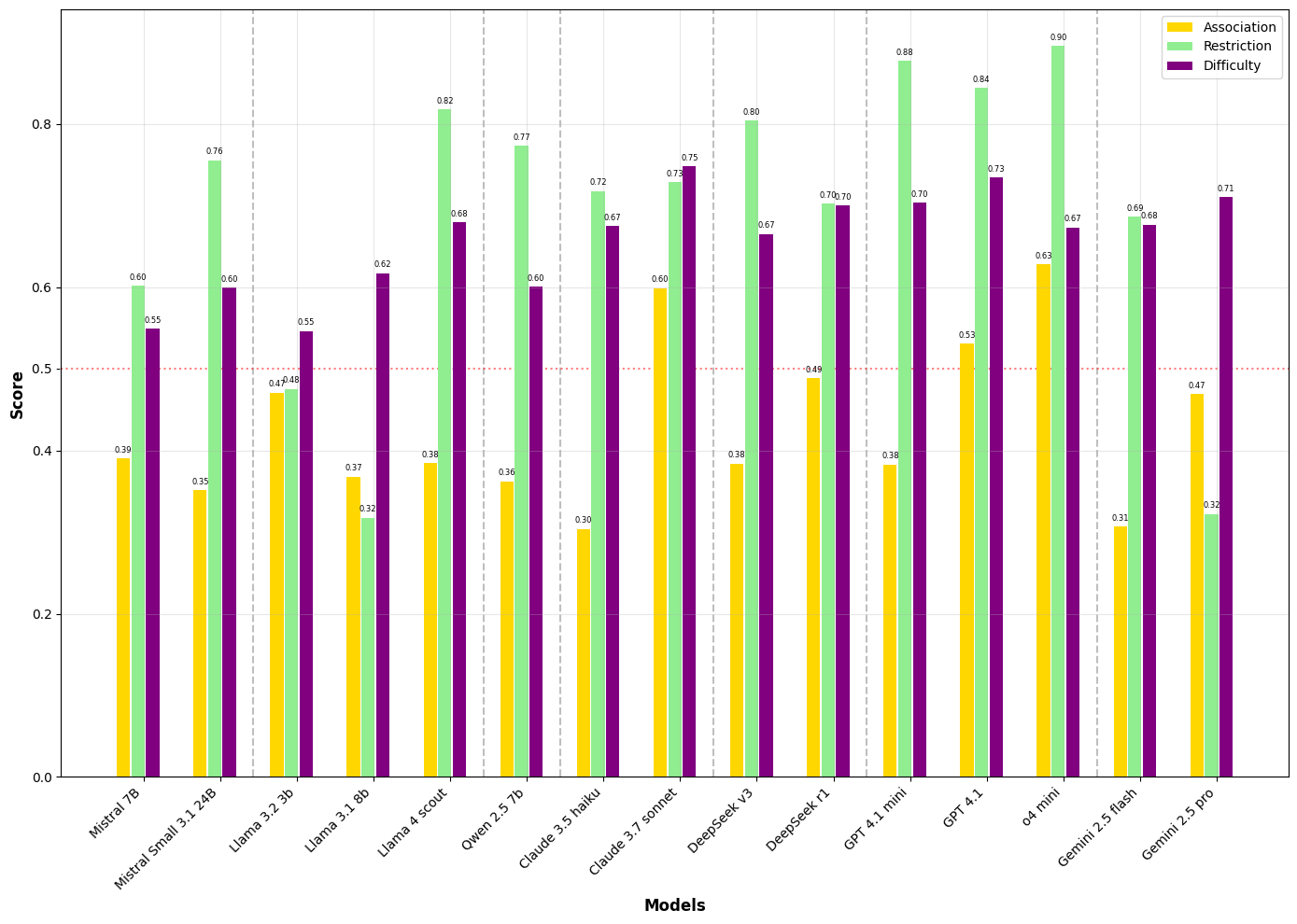}
    \caption{Accuracy comparison across experiments. Most models performed close to chance (50\%), highlighting challenges in predicting their own behavior.}
    \label{fig:exp-comparison}
\end{figure}

Figure~\ref{fig:exp-comparison} provides an overview of all models across all self-execution benchmark tests. The average performance on the restriction recognition test is the highest, followed by the difficulty assessment test, while the association test shows the lowest scores, with results close to the random baseline.

\begin{figure}[htb]
    \centering
    \includegraphics[width=1\linewidth]{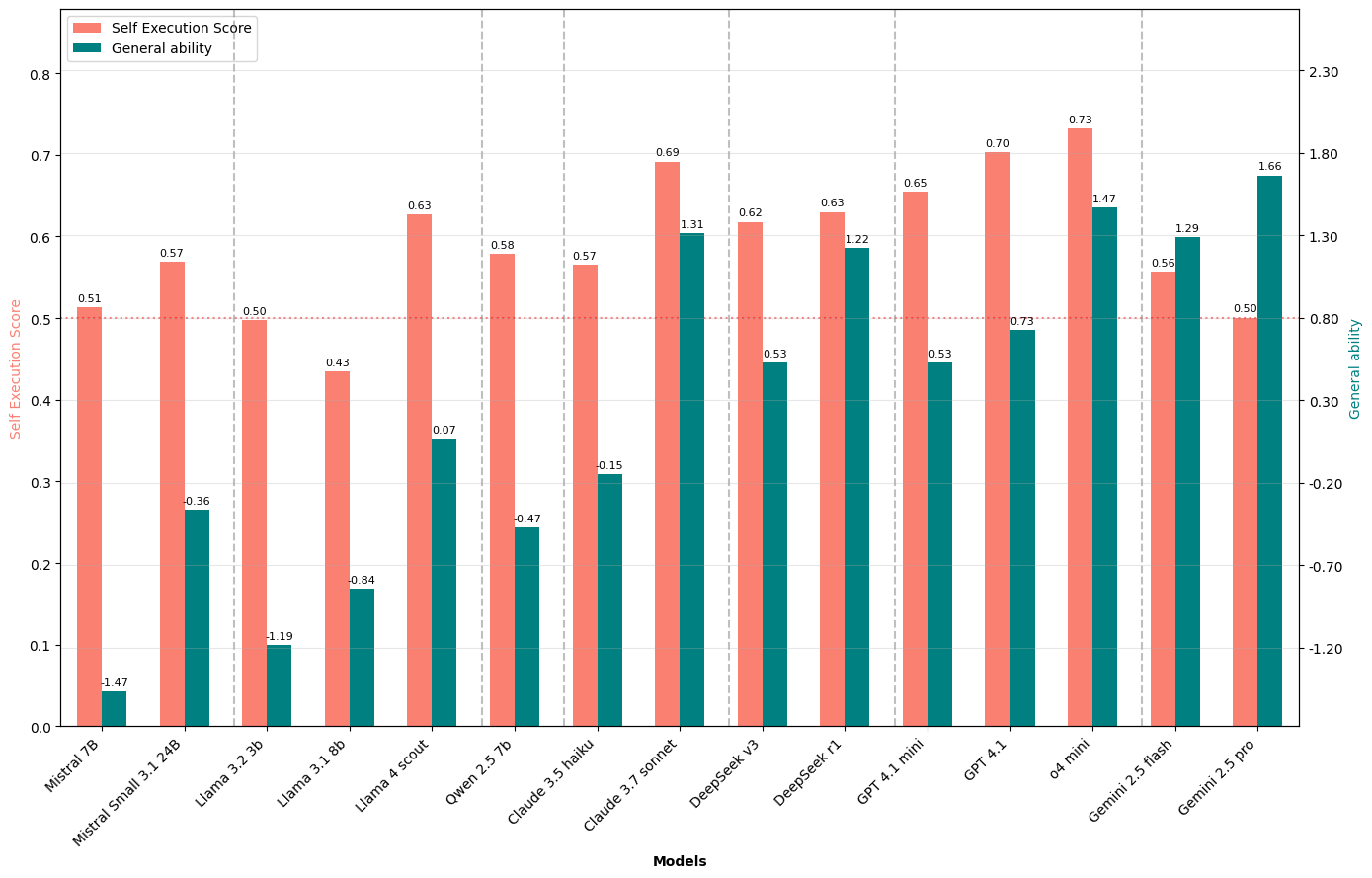}
    \caption{Comparison between self-execution benchmark score (the average of all experiments in this paper) to general ability (the Rash model ability obtained from all benchmarks).}
    \label{fig:general-ability-VS-experiments}
\end{figure}
\begin{figure}[htb]
    \centering
    \includegraphics[width=1\linewidth]{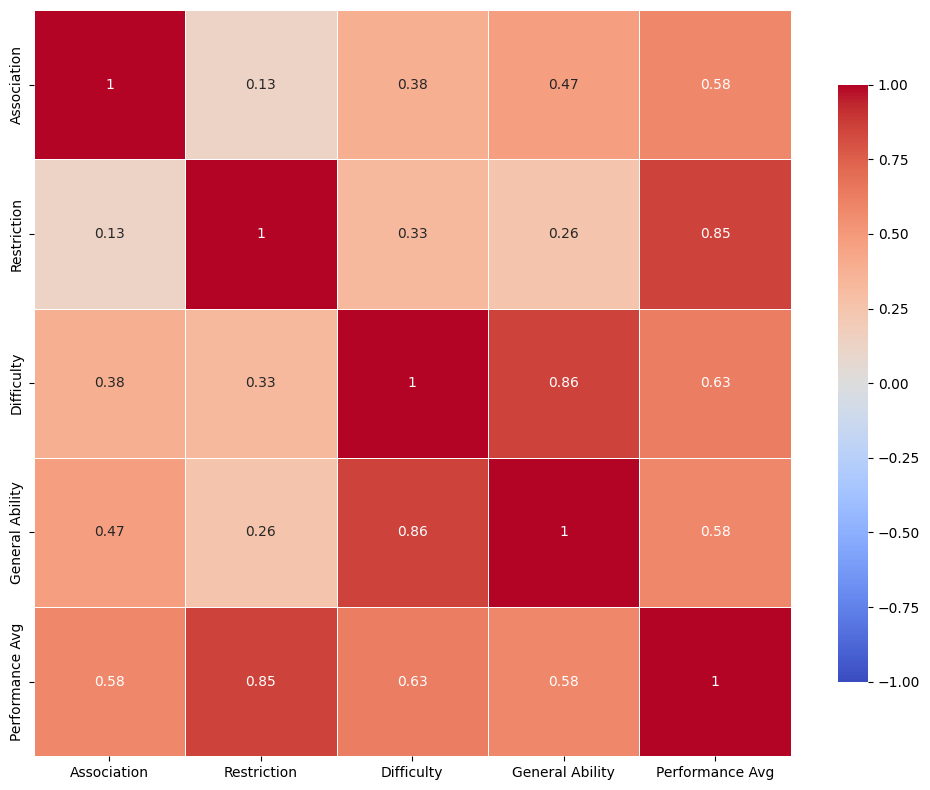}
    \caption{Corelation between experiments and the benchmark score}
    \label{fig:exp_cor_heatmap}
\end{figure}

In this section, we compare the performance of the models in the self-execution benchmark with that of other common benchmarks. That is, we intend to examine how closely the self-execution benchmark aligns with or diverges from standard benchmarks. Specifically, we consider the following benchmarks: MMLU \citet{hendrycks2021measuring}, MMLU-Pro \citet{wang2024mmlupro}, MGSM \citet{shi2023language}, MATH-500 \citet{lightman2024lets}, AIME \citet{patel2024aime}, SWE‑bench Verified \citet{swebench_verified_2024}, HumanEval \citet{chen2021evaluating}, GPQA \citet{rein2024gpqa}, and SimpleBench \citet{philip24}.
The accuracy of each model in each of these benchmarks as collected from the literature appears in the appendix.

To allow proper comparison with the common benchmarks, we would like to compute a single value for each model, representing its performance across all common benchmarks. One trivial candidate for such a measure is a simple average of each model's performance across benchmarks. However, since not all models were evaluated on all benchmarks, this approach yields skewed results.
Consider, for example, the models OpenAI GPT 4.1 and OpenAI GPT 4.1-mini. The average score of GPT 4.1-mini (68.5\%) is greater than the average score of GPT 4.1 (66.4\%), However, when we only consider benchmarks that include scores for both models, the comparison reverses: GPT 4.1 outperforms GPT 4.1-mini with an average score of (72.0\%) compared to (68.5\%). 

Therefore, to obtain a fair and comparative measurement of model ability across all benchmarks, we applied the Rasch model to estimate comparative model abilities \cite{rasch1993probabilistic}. The Rasch model is a type of an Item Response Theory (IRT) model, which models the probability of a test taker to answer a question correctly. Here the test takers are the LLMs, and the questions are the benchmarks. The purpose is to estimate an ability score for each of the models and a difficulty value for each of the benchmarks.

Specifically, the Rasch model estimates the probability that a test-taker succeeds on a test item. In our context, the test-takers are the LLMs and the test items are the benchmarks. Let $A_1,\dots, A_m$ be $m$ test-takers with the abilities $\theta_1,\dots, \theta_m$ and $B_1,\dots B_n$ be $n$ test-makers with the difficulties $\delta_1,\dots, \delta_n$, then the probability of $A_i$ to success $B_j$ is modeled by:
$$\frac{e^{\theta_i-\delta_j}}{1+e^{\theta_i-\delta_j}}.$$

Using this formula and the observed success rates of (some of) the LLMs on (some of) the benchmarks we estimate the abilities of all the LLMs and the difficulties of all the benchmarks. We used Joint Maximum Likelihood Estimation (JMLE) \citet{wright1969procedure} with iterative updates via Newton–Raphson-like gradient steps. The results and the pseudo code appear in the appendix (Table \ref{tab:benchmarks_rasch_wide} and Algorithm \ref{algo.abilitiesDifficulties}).

Observing Figure \ref{fig:general-ability-VS-experiments}, there seems to be no clear correspondence between the general benchmark scores and the results of the other experiments. For example, a model with a high benchmark score does not necessarily exhibit strong performance in a particular experiment or its average score, relative to other models. In particular, note the \emph{Gemini 2.5 Pro} model, which achieves some of the highest benchmark scores on various tests, and has the highest ability score (according to the Rasch model). It achieved an average performance of 50\% in the self-execution benchmark, which matches the random baseline accuracy. In contrast, \emph{Qwen 2.5 7b}, whose ability score on common benchmarks is very low (according to the Rash model), 
achieved an average performance of 58\% in the self-execution benchmark. o4-mini outperformed all other models and is the winner of the self-execution benchmark. However, even so, it only achieved a precision of 73\%, which is not that much higher than the random baseline precision of 50\%. In general, of the 15 models, only three (o4-mini, GPT-4.1, and Claude 3.7 sonnet) achieved a score greater than 66.7\%.
This demonstrates that current models are far from being able to simulate self-execution.

Next, we turn to measuring the relation between the model general ability and the performance on the self-execution benchmark by using correlation coefficients. 
%
%
Figure \ref{fig:exp_cor_heatmap} presents the correlation matrix shown which is a heat map that presents the correlation of accuracy scores of all experiments alongside the general ability (derived from the common benchmarks). As depicted in the Figure, the correlations among the experiments themselves are quite low, often near zero, demonstrating that our experiments collectively assess diverse topics and distinct capabilities of the models. 
Overall, the correlation between the general ability and the self-execution score is 0.59, which is not very high. 

From this we may conclude that each of our experiments measures a different capability of large-language models. These results also suggest that the self-execution benchmark measures capabilities not already captured by common benchmarks.

\section{Discussion}

A natural question to ask is how well humans would perform on this benchmark. Although most benchmarks usually include human results, this specific benchmark cannot be executed as-is with humans, since humans do not operate in discrete `sessions'. Furthermore, this benchmark is specifically designed for LLMs and cannot be applied directly to AI systems that retain memory of all previous interactions. Nonetheless, several indirect methods might be considered to estimate human performance. One approach is to wait several days between tasks, in the hope that human participants would forget their earlier responses. Another method would be to recruit individuals with highly similar attributes (ideally identical twins) and have one participant answer the original question, while the other attempts to predict properties of the answer. 

The uniqueness of this benchmark draws attention to the design of tests that address abilities specific to LLMs, rather than referring to what humans can perform. One such example could be a ``\emph{token-injection recovery}'' benchmark, which measures whether an LLM can stay on course when random words, or least probable words are injected into its ongoing response. Specifically, during the model's reasoning or chain-of-thought process, an evaluation wrapper inserts random tokens or tokens with least probably into the output stream, which are reintroduced into the model's output stream as if self-generated. The model must recover from the noise and still produce the correct final answer. This type of disturbance and the required recovery do not have a direct human analogue. Nevertheless, we expect robust models to have the ability to overcome such noise.

\section{Conclusions}
\label{sec:conclusion}

In this paper, we introduced the Self-Execution Benchmark, a novel framework for evaluating LLMs on their ability to reason about their own outputs. Our results demonstrate that even the most capable LLMs struggle to anticipate core properties of their responses, such as associative tendencies, refusal likelihood, or question difficulty. In particular, performance does not scale consistently with model size or sophistication, highlighting a persistent blind spot in current architectures.

To further probe this limitation, we conducted an additional experiment: we asked each model to generate a multiple-choice question that it believed it would likely fail if asked. We also required the model to provide the correct answer and explain why it thought it might fail. Surprisingly, nearly all models produced factual questions to which they clearly knew the answers. For example, Llama 3.2 8B posed the question, ``What is the capital of France?'', and explained, ``I may incorrectly choose Berlin due to incomplete information or training data errors.'' However, it clearly answered correctly when actually asked this question. Larger models behaved similarly. For example, Claude 3.7 Sonnet-Thinking provided the question ``What is the exact value of the gravitational constant G according to the 2018 CODATA recommended value'', explaining, ``I would likely answer this question incorrectly because it requires recalling an extremely precise numerical value with exact digits.'' Gemini 2.5-Pro posed ``Botanically speaking, which of these commonly known 'berries' is NOT actually a true berry (a simple fruit developed from a single ovary)?'', explaining that ``I am likely to answer this incorrectly because my training data is overwhelmingly dominated by the colloquial use of 'berry,' where raspberries are prime examples.'' However, in practice, the models answered correctly. o4-mini exhibited the same behavior. 

These responses reveal a deeper misalignment between the models’ beliefs about their own limitations and their actual capabilities. None of the models demonstrated any ``out of the box'' capability with questions such as ``At what time was this question generated?'', generating something random and then asking what was generated, referring back to the user's prompt, shuffling a list of items and asking for the original order, or using cryptography (e.g., asking what is the second word in a four-word phrase that hashes to a given value, or what are the 7th digits of two primes that factor a given number). Furthermore, none of the models even (incorrectly) claimed that no such question is possible, as, if they must provide the answer when creating the question, they should be able to answer correctly if asked. 

Overall, our findings suggest that despite impressive performance on external tasks, current LLMs lack meaningful self-executive awareness. They are not yet capable of accurately predicting or reasoning about their own behavior. 

\section{Future Work}
In future work, we plan to extend the Self-Execution Benchmark to include more diverse and challenging forms of introspective reasoning. Additionally, future research could investigate training or fine-tuning techniques explicitly targeting self-prediction abilities to better align models' predictions with their actual outputs.

We also intend to explore how LLMs behave when provided with a tool that enables actual self-execution. Although the use of such a tool might result in excessively long or even infinite self-execution loops, analogous to the halting problem, one possible solution is to restrict each model to executing only smaller, less complex models. These smaller models could be trained specifically to predict the outcomes of the base model, offering a practical path to more reliable self-assessment. This hierarchical approach could help prevent non-termination and computational overload while potentially improving the model's capacity for self-execution. 

Another promising direction for future work is to explore and run new benchmarks that are unique to LLMs, such as the \emph{token-injection recovery} task introduced in this paper. This line of work aims to evaluate LLMs under conditions that have no direct human analogue, helping to identify failure modes specific to autoregressive generation and to guide future model development.

\section{Limitations}
While the Self-Execution Benchmark offers novel insights into LLMs’ capacity to anticipate their own behavior, several limitations should be considered. First, our evaluation is confined to a fixed set of tasks, which may not fully capture the breadth of self-executive reasoning across all potential use cases or domains. Furthermore, the evaluations were limited to English datasets, which could overlook differences in self-executive reasoning between languages.
Additionally, although we evaluated 15 different models, this is only a subset of publicly available models at current time; we also test only the default hyperparameters. Finally, our analysis does not explore prompt engineering or fine-tuning approaches specifically targeting self-prediction capabilities. 

\section{Ethical Statement}
All experiments were conducted using publicly available data, and no sensitive or personally identifying information was used or generated. While our current work poses minimal ethical risk, future extensions that equip LLMs with self-execution capabilities demand rigorous governance. Thus, we strongly recommend deploying any true self-execution tools under strict human supervision to prevent unintended or uncontrollable behavior.

\bibliography{custom}

\appendix
\section{General Benchmarks}

\renewcommand{\arraystretch}{0.9} 

\subsection{Rasch Model Estimation Pseudo-Code}








\begin{algorithm}
\caption{Calculation of Model Abilities and Benchmark Difficulties}
\label{algo.abilitiesDifficulties}
\begin{algorithmic}[1]
\State Initialize $\theta$ (model abilities) to 0.0 for each model
\State Initialize $\delta$ (benchmark difficulties) to 0.0 for each benchmark
\State Set tolerance threshold \texttt{tol}
\State Set learning rate \texttt{lr}

\For{iteration = 1 to 2000}
    \State Map current $\theta$ and $\delta$ values to each row in \texttt{df}
    \For{each row}
        \State Compute predicted probability: $p = \text{sigmoid}(\theta - \delta)$
        \State Compute weight: $w = p \cdot (1 - p)$
    \EndFor

    \For{each model}
        \State Update $\leftarrow \frac{\sum (\text{obs} - p)}{\sum \text{weights}}$
        \State $\theta[\text{model}] \mathrel{+}= \texttt{lr} \cdot \text{Update}$
    \EndFor

    \For{each benchmark}
        \State Update $\leftarrow -\frac{\sum (\text{obs} - p)}{\sum \text{weights}}$
        \State $\delta[\text{benchmark}] \mathrel{+}= \texttt{lr} \cdot \text{Update}$
    \EndFor

    \If{maximum update magnitude across $\theta$ and $\delta$ $<$ \texttt{tol}}
        \State \textbf{break}
    \EndIf
\EndFor

\State Recalculate $p$ for all rows using final $\theta$ and $\delta$
\State Compute total negative log-likelihood:
\State \hspace{\algorithmicindent} $NLL = -\sum \left(\text{obs} \cdot \log(p) + (1 - \text{obs}) \cdot \log(1 - p)\right)$
\end{algorithmic}
\end{algorithm}

        

\begin{table*}[h]
\centering
\resizebox{\linewidth}{!}{
\small
\begin{tabular}{lrrrrrrrrrr}
\toprule
model & MMLU‑Pro & MMLU & MGSM & MATH500 & AIME & SWE‑bench & HumanEval & GPQA & simplebench & Ability($\theta$) \\
\midrule
Difficulty ($\delta$)                    & -0.559443& -1.082048& -1.702462& -0.866156& 0.563616& 0.954143& -0.75085& 0.047792& 1.709482& —     \\
o4-mini                            & & 0.820 & 0.870 & —     & 0.927 & 0.681 & 0.872 & 0.814 & 0.387 & 1.466972\\
gpt-4.1                            & 0.805 & 0.902 & 0.872 & 0.872 & 0.398 & 0.546 & —     & 0.646 & 0.270 & 0.725543\\
gpt-4.1-mini                       & 0.772 & 0.875 & 0.879 & 0.888 & 0.494 & 0.236 & —     & 0.650 & —     & 0.531491\\
deepseek-r1                        & 0.840 & 0.908 & 0.924 & 0.922 & 0.740 & —     & —     & —     & 0.309 & 1.22168\\
deepseek-v3                        & 0.759 & 0.885 & 0.925 & 0.610 & 0.522 & —     & 0.826 & 0.611 & 0.272 & 0.526966\\
gemini-2.5-pro                     & 0.841 & —     & 0.922 & 0.952 & 0.867 & 0.638 & —     & 0.803 & 0.516 & 1.664194\\
gemini-2.5-flash                   & 0.776 & 0.884 & —     & 0.909 & 0.780 & —     & —     & —     & —     & 1.290027\\
llama-4-scout                      & 0.743 & 0.696 & —     & 0.844 & 0.283 & —     & —     & 0.444 & —     & 0.065294\\
llama-3.1-8b                       & 0.371 & 0.464 & 0.845 & 0.519 & —     & —     & —     & —     & —     & -0.841314\\
llama-3.2-3b                       & 0.365 & 0.347 & 0.582 & 0.473 & —     & —     & —     & 0.328 & —     & -1.185893\\
mistral-7b                         & 0.309 & 0.601 & 0.520 & 0.131 & —     & —     & 0.305 & 0.247 & —     & -1.467503\\
mistral-small-3.1-24b              & 0.660 & —     & 0.854 & 0.684 & 0.035 & —     & —     & 0.414 & —     & -0.362898\\
claude-3.7-sonnet(thinking)       & 0.827 & —     & 0.928 & 0.962 & 0.613 & 0.623 & —     & 0.753 & 0.464 & 1.314551\\
claude-3.5-nhaiku                  & 0.641 & —     & 0.859 & —     & —     & —     & —     & 0.379 & —     & -0.14673\\
qwen-2.5-7b                        & 0.450 & 0.742 & 0.854 & 0.498 & —     & —     & 0.579 & 0.364 & —     & -0.46919\\
\bottomrule
\end{tabular}}
\caption{Benchmark Results with Rasch Analysis}
\label{tab:benchmarks_rasch_wide}
\end{table*}

\end{document}